\documentclass[11pt]{article}

\usepackage[preprint]{acl}

\usepackage{times}
\usepackage{latexsym}
\usepackage{float}
\usepackage{microtype}
\usepackage{graphicx}
\usepackage{subcaption}
\usepackage{booktabs} 
\usepackage{amsfonts}
\usepackage{amsmath}
\usepackage{nicefrac}
\usepackage{microtype}
\usepackage{xcolor}
\usepackage{graphicx}
\usepackage{multirow}
\usepackage{natbib}
\usepackage{comment}
\usepackage{xcolor}
\usepackage{tikz}
\usepackage{pgfplots}
\pgfplotsset{compat=1.18}
\usepackage{multirow}
\usepackage{microtype}
\usepackage{caption}
\usepackage{subcaption}
\usepackage{colortbl}
\usepackage{tikz}
\usepackage{pgfplots}
\usepackage{tikz}
\usepackage{pgfplots}
\usepackage{CJKutf8} 
 
\pgfplotsset{compat=1.18}
\usepackage{multirow}
\usepackage{microtype}
\usepackage{caption}
\usepackage{subcaption}
\usepackage{colortbl}
\usepackage{tikz}
\usepackage{pgfplots}
 
\pgfplotsset{compat=1.18}
\usepackage{array}
\usetikzlibrary{
  arrows.meta, positioning, shapes.geometric,
  fit, backgrounds, calc, decorations.pathreplacing
}
\usepackage{tcolorbox}
\tcbuselibrary{skins,breakable}
\tcbset{
  promptbox/.style={
    enhanced, breakable,
    colback=blue!5, colframe=blue!35,
    fontupper=\small\ttfamily,
    boxrule=0.4pt, arc=3pt,
    left=6pt, right=6pt, top=4pt, bottom=4pt,
  },
  prompttitled/.style={
    promptbox, coltitle=black, colbacktitle=blue!18,
    fonttitle=\bfseries\small, attach boxed title to top left={yshift=-2mm, xshift=4mm},
    boxed title style={colframe=blue!35, boxrule=0.4pt, arc=2pt}
  }
}

\definecolor{cright}{RGB}{34,139,34}
\definecolor{cwrong}{RGB}{178,34,34}
\definecolor{cmlp}{RGB}{200,60,60}
\definecolor{cattn}{RGB}{60,160,60}
\definecolor{cblue}{RGB}{70,130,180}
\definecolor{cgray}{RGB}{160,160,160}
\definecolor{corange}{RGB}{220,120,30}

\usepackage[T1]{fontenc}

\usepackage[utf8]{inputenc}

\usepackage{microtype}

\usepackage{inconsolata}

\usepackage{graphicx}

\title{Repeated-Token Counting Reveals a Dissociation Between \\ Representations and Outputs}

\author{Sohan Venkatesh \\
  Manipal Institute of Technology Bengaluru }

\begin{document}
\maketitle
\begin{abstract}
Large language models fail at counting how many times a word repeats in a list, even though they perform well on far harder reasoning tasks. These failures are commonly attributed to limitations in internal count tracking. We show this attribution is wrong. Linear probes on the residual stream decode the correct count with near-perfect accuracy at every post-embedding layer and they do so even at the exact layers where the wrong answer crystallizes in the output. Attention patterns show no evidence of collapse over repeated tokens and tokenization artifacts account for none of the failure. Instead, a multi-layer perceptron (MLP) block at roughly 85--93\% network depth overwrites the correctly-encoded count with a fixed wrong answer. Ablating this block changes the wrong output and establishes it as causally responsible for the failure. The block fires on the space-separated repeated-word format and is absent for repeated digit-tokens. The pattern holds across Llama-3.2 (1B and 3B) and Qwen2.5 (1.5B, 3B and 7B) at consistent relative depth. The count is represented correctly and a specific computation prevents it from reaching the output, so representation failures and routing failures require different interventions.
\end{abstract}

\section{Introduction}
\label{sec:intro}
 
Counting repeated tokens is one of the simplest list-processing tasks and large language models fail at it in ways that are systematic and reproducible. Ask a model to count ten repeated words and it gives a confident wrong answer with zero variance across seeds under greedy decoding. In some models, changing the delimiter from spaces to commas rescues the output.
In others, the wrong answer persists regardless of format. Both outcomes challenge the common association of counting failures with the lack of internal counting representations.
 
\begin{figure*}[t]
\centering
\includegraphics[width=\linewidth, trim=8 7 8 42, clip]{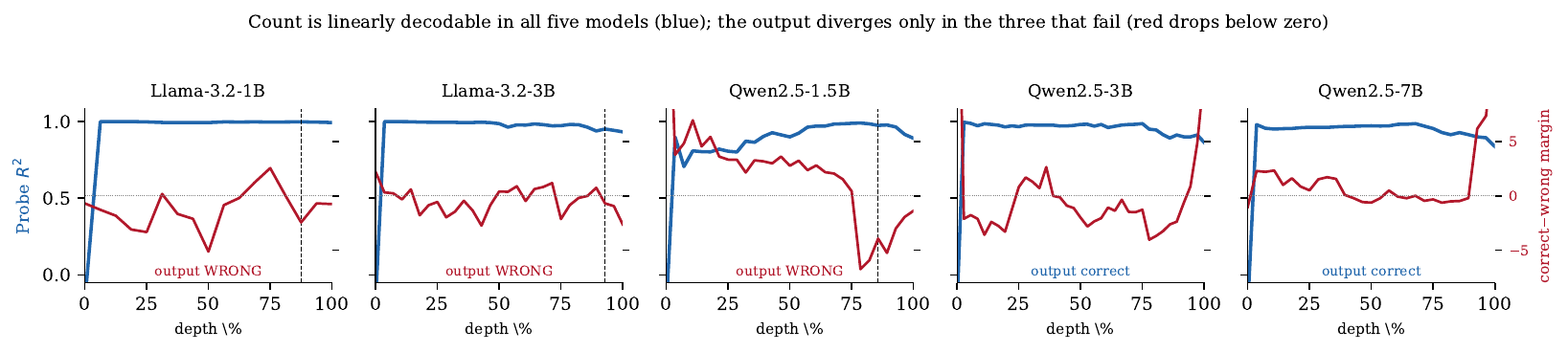}
\caption{\textbf{The count is linearly decodable in all five models but reaches the output only in the models that solve counting.} Each panel is one model. The blue curve gives the linear-probe $R^2$ for the repeated-token count at each relative network depth and the red curve gives the correct-minus-wrong logit margin at the output projection. In the three counting-failure models (Llama-3.2-1B, Llama-3.2-3B and Qwen2.5-1.5B) the red curve drops below zero at the writer layer (dashed) while the blue curve stays high. This divergence is the dissociation, with the count present in the residual stream while the output commits to the wrong answer. In the two models that solve counting (Qwen2.5-3B and Qwen2.5-7B), both curves stay high, so the decodable count reaches the output.}
\label{fig:intro}
\end{figure*}
 
The format does not contain the answer. A model that succeeds under comma format and fails under space format has the correct count internally under both conditions.
A model that fails under both formats may have the count as well. The question is not whether the count is represented but whether a specific computation prevents it from reaching the output. A representation failure calls for better training or architecture. A routing failure calls for identifying which circuit intercepts the signal and why it fires. Conflating the two leads to interventions that address the wrong problem. Figure~\ref{fig:intro} summarizes this representation-output dissociation across all five models.
 
We study repeated-token counting as a tractable test case for this distinction. The task is simple enough that the correct answer is a single integer provable at every layer, yet the failure is systematic and deterministic. We find a representation-behavior dissociation in Llama-3.2, where the count is linearly decodable at the layer at which the wrong answer crystallizes in the output. The responsible MLP block receives count-invariant input and produces a fixed wrong output regardless of actual sequence length. The prior is word-type-specific, firing for repeated alphabetic tokens but not digit-tokens and it is disengaged by any delimiter that breaks the contiguous-space format.
 
We replicate this pattern in Qwen2.5 and show that in the larger models that solve counting, anomaly detection fails for the same reason. The relevant signal is present in attention weights but does not reach the
output. The behavioral failures in capable models may more often be failures of expression than failures of representation and identifying which type of failure is present is the necessary first step before any intervention can succeed.

\section{Related Work}
\label{sec:related}
 
\paragraph{Counting in language models.}
\citet{zhang2024counting} document counting failures across token types and list lengths and show that tokenization artifacts explain only a fraction of errors. \citet{datta2026early} use linear probes and logit lens to study character-level counting in Llama and Qwen, finding the correct count is encoded early but suppressed at later layers. We study repeated-token counting over word lists instead of the character-level counting they examine. \citet{hasani2026mechanistic} show through attention analysis that counting ability is bounded by transformer depth. \citet{levy2024same} show that input length degrades LLM reasoning across diverse tasks. \citet{shi2023large} show irrelevant context causes numerical reasoning failures despite correct intermediate steps. \citet{anil2022exploring} and \citet{wei2022chain} show that chain-of-thought prompting improves numerical performance by making intermediate steps explicit.

\paragraph{Mechanistic interpretability.} \citet{elhage2021mathematical} introduce the residual stream framework and the mathematical tools for decomposing transformer computations into per-layer contributions.
\citet{meng2022locating} show that factual associations are stored in MLP layers at mid-to-late network depth using causal mediation analysis. \citet{stolfo2023mechanistic} use causal mediation to trace arithmetic reasoning and find that late-layer MLP modules write result-related information into the residual stream.

\paragraph{Representation versus routing.}
\citet{marks2023geometry} show that language models linearly encode the truth or falsehood of factual statements even when their outputs are wrong, using linear probes and causal interventions on residual stream activations. \citet{burns2022discovering} develop an unsupervised method to elicit latent knowledge from models that produce incorrect outputs. \citet{tigges2023linear} show linear sentiment representations exist in model internals independently of behavioral output. This body of work establishes that a gap between internal representations and model outputs is a general phenomenon. Our contribution is a circuit-level mechanism for one instance of this gap. We identify the specific MLP block that overwrites the correct count and the format signature that triggers it.
 
\paragraph{MLP layers as memory.}
\citet{geva2021transformer} show that MLP layers function as key-value memories where keys are input patterns and values are output distributions. \citet{geva2022transformer} extend this by showing that specific MLP layers promote particular concept tokens in the residual stream during a forward pass. Our finding that a single MLP block writes a fixed wrong answer for repeated word-token inputs is a direct instance of this mechanism.
 
\paragraph{Format sensitivity and attention.}
\citet{xiao2024efficient} identify attention sink tokens that receive disproportionate attention mass from the last position regardless of content. \citet{press2023measuring} show that prompt format substantially affects performance and that models are sensitive to surface-level phrasing.

\section{Experimental Setup}
\label{sec:setup}
 
\subsection{Models}
 
We study five instruction-tuned models: Llama-3.2-1B-Instruct, Llama-3.2-3B-Instruct \citep{meta2024llama32}, Qwen2.5-1.5B-Instruct, Qwen2.5-3B-Instruct and Qwen2.5-7B-Instruct \citep{yang2024qwen25}. The two Llama models share architecture, tokenizer, chat template and training recipe, varying only in scale, which controls for confounds that arise when comparing model families. All inference uses greedy decoding (temperature=0). Each condition is run over ten seeds and produces identical outputs, so accuracy per condition is binary. We use a separately loaded model with \texttt{attn\_implementation="eager"} in mechanistic analysis to enable attention weight extraction.
 
\subsection{Task and Prompt Design}
 
Three conditions each present a list of ten tokens and request a count.
\begin{enumerate}
  \item \textbf{P1 Baseline.} Ten identical tokens. Count that token. The correct answer is 10.
  \item \textbf{P2 Anomaly.} Nine of the repeated token and one intruder
  (``banana'') at position~5. Count the repeated token. The correct answer is 9.
  \item \textbf{P3 Control.} Ten unique words. Count the words. The correct answer is 10.
\end{enumerate}
The P1 prompt template is:
\begin{tcolorbox}[promptbox]
Count the number of times "apple" appears in this list: apple apple apple
apple apple apple apple apple apple apple. Respond only with the integer,
nothing else.
\end{tcolorbox}
P2 replaces one apple with banana at position~5. P3 replaces all tokens with distinct words. Each condition is also run with comma-separated delimiters. Those prompts use the phrase ``in this list'' without the word ``comma-separated'' in the instruction to avoid introducing a format confound. The full set of conditions, format variants, paraphrases and token sets is reproduced in Appendix~\ref{sec:prompts}.
 
\subsection{Mechanistic Analysis Methods}
\label{sec:methods}
 
We tokenize the word-list payload of each prompt to verify word count equals token count. We then run a single forward pass per phase with attention output enabled and extract last-token attention at each layer restricted to word-list positions, masking out the beginning-of-sequence (BOS) token and prompt tokens. The per-layer entropy and uniformity over this span are computed for all three phases to test the attention-sink hypothesis.
 
For linear probes we generate 13 prompts with $n\in[3,15]$ repeated tokens and 11 prompts with $n\in[3,13]$ unique words, extract the last-token residual stream at every layer and train a Ridge regression probe per layer using leave-one-out cross-validation to predict $n$.
We report mean absolute error and $R^2$ per layer for both conditions separately \citep{belinkov2022probing}. For the logit lens we project the last-token hidden state through the final layernorm and unembedding matrix at each layer, identify the top predicted digit token (integers 1--19, single-token only) and normalize layer indices to fraction of total depth for cross-model comparison \citep{nostalgebraist2020logitlens}.
 
For MLP vs.\ attention decomposition we extract three residual stream states per layer, namely $\mathbf{h}_\text{before}$ entering the decoder layer, $\mathbf{h}_\text{post-attn}$ after the attention sublayer and before the MLP and $\mathbf{h}_\text{post-layer}$ after the full layer. We write $d_0$, $d_1$ and $d_2$ for the top logit-lens digit read from each. Each state is projected through the logit lens and the component introducing the wrong answer is classified as MLP, ATTENTION, MLP~ERASES or stable \citep{vig2020causal,goldowsky2023localizing}. We run this decomposition at layers surrounding the logit lens lock-in point (L11--L16 for 1B, L22--L28 for 3B) and additionally sweep the identified writer layer across $n\in\{7,8,9,10,11,12,15\}$ to test count-invariance. We zero the primary MLP writer output to test causal involvement.
 
Paraphrase analysis applies logit lens and MLP decomposition to five surface-form variations of the P1 instruction (original, how\_many, list\_first, tally, simple) recording lock-in layer, depth and writer label per paraphrase. Symbol substitution replaces ``apple'' with symbols from $\{$apple, cat, the, a, X, 1, 0, 7$\}$ at $n{=}10$ to test word-type specificity of the prior.
For Qwen2.5-3B and 7B we vary banana position (0--9) and banana count (1--5) to characterize anomaly detection position-dependence and quantity-threshold and compute per-layer banana-to-apple attention ratios with per-head analysis at the most banana-ignoring layers.

\section{Behavioral Results}
\label{sec:behavioral}
 
\subsection{Counting failure is format-specific and scale-dependent}
Table~\ref{tab:behavioral} shows behavioral accuracy across all five models and both formats. Every failure is deterministic, with identical outputs across all ten seeds. All payloads tokenize to exactly 10 tokens under space-separated format across all the models which rules out byte-pair encoding (BPE) merging artifacts as a confounder.
 
\begin{table*}[t]
\centering
\caption{Accuracy under space-separated (SP) and comma-separated (CS) format
across all five models. 0\,\% cells show the attractor value (deterministic wrong answer) in parentheses. Type C means the model fails P1 (cannot count repeated tokens). Type A means the model solves P1 but fails P2 (cannot detect the anomaly).}
\label{tab:behavioral}
\setlength{\tabcolsep}{3pt}
\small
\begin{tabular}{lcccccc}
\toprule
\textbf{Model} & \textbf{P1 SP} & \textbf{P1 CS} & \textbf{P2 SP} &
\textbf{P3 SP} & \textbf{P3 CS} & \textbf{Type} \\
\midrule
Llama-1B
  & \textcolor{cwrong}{0\,\% (``8'')}
  & \textcolor{cwrong}{0\,\% (``8'')}
  & \textcolor{cwrong}{0\,\% (``8'')}
  & \textcolor{cright}{100\,\%}
  & \textcolor{cwrong}{0\,\% (``14'')}
  & C \\
Llama-3B
  & \textcolor{cwrong}{0\,\% (``14'')}
  & \textcolor{cright}{100\,\%}
  & \textcolor{cwrong}{0\,\% (``8'')}
  & \textcolor{cwrong}{0\,\% (``11'')}
  & \textcolor{cwrong}{0\,\% (``11'')}
  & C \\
Qwen-1.5B
  & \textcolor{cwrong}{0\,\% (``8'')}
  & \textcolor{cright}{100\,\%}
  & \textcolor{cwrong}{0\,\% (``6'')}
  & \textcolor{cright}{100\,\%}
  & \textcolor{cright}{100\,\%}
  & C \\
Qwen-3B
  & \textcolor{cright}{100\,\%}
  & \textcolor{cright}{100\,\%}
  & \textcolor{cwrong}{0\,\% (``10'')}
  & \textcolor{cright}{100\,\%}
  & \textcolor{cright}{100\,\%}
  & A \\
Qwen-7B
  & \textcolor{cright}{100\,\%}
  & \textcolor{cright}{100\,\%}
  & \textcolor{cwrong}{0\,\% (``8'')}
  & \textcolor{cright}{100\,\%}
  & \textcolor{cright}{100\,\%}
  & A \\
\bottomrule
\end{tabular}
\end{table*}
 
Three models fail P1 under space format while Qwen2.5-3B and 7B solve it completely. This places the capability threshold at Qwen2.5-3B, with the Llama family and Qwen2.5-1.5B retaining the counting prior.
 
Format sensitivity differs across models. In Llama-3.2-1B the prior is format-invariant, since comma format leaves P1 and P2 unchanged while activating a separate prior that collapses P3 to a deterministic ``14''. Comma format rescues P1 in both Llama-3.2-3B and Qwen2.5-1.5B but restores P2 only in 3B. Under space format, by contrast, P2 fails in every model regardless of scale or family.

\subsection{Attractor structure reveals model-specific priors}

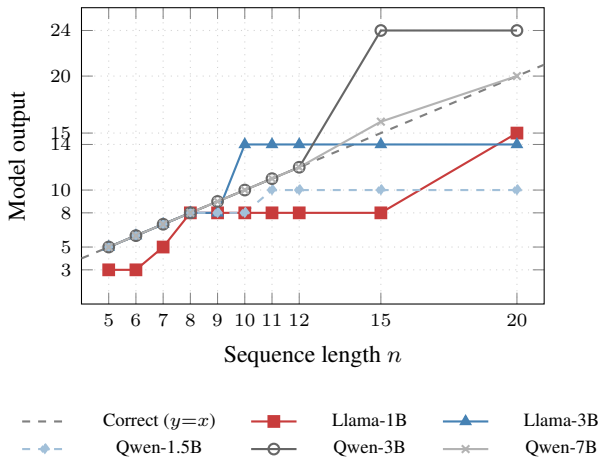
\begin{figure}[H]
\centering
\begin{tikzpicture}
\begin{axis}[
  width=\columnwidth,
  height=5.5cm,
  xlabel={Sequence length $n$},
  ylabel={Model output},
  xmin=4, xmax=21,
  ymin=0, ymax=26,
  xtick={5,6,7,8,9,10,11,12,15,20},
  ytick={3,5,8,10,14,15,20,24},
  grid=major,
  grid style={dotted,gray!35},
  tick label style={font=\scriptsize},
  label style={font=\small},
  legend columns=3,
  legend style={
    font=\scriptsize,
    inner sep=2pt,
    cells={align=left},
    at={(0.5,-0.33)},
    anchor=north,
    draw=none,
    column sep=1em,
  },
]
 
\addplot[dashed,gray,thick,domain=4:21,samples=2]{x};
\addlegendentry{Correct ($y{=}x$)}
 
\addplot[cmlp,mark=square*,thick,mark size=2pt] coordinates {
  (5,3)(6,3)(7,5)(8,8)(9,8)(10,8)(11,8)(12,8)(15,8)(20,15)};
\addlegendentry{Llama-1B}
 
\addplot[cblue,mark=triangle*,thick,mark size=2pt] coordinates {
  (5,5)(6,6)(7,7)(8,8)(9,8)(10,14)(11,14)(12,14)(15,14)(20,14)};
\addlegendentry{Llama-3B}
 
\addplot[cblue!50,mark=diamond*,thick,mark size=2pt,dashed] coordinates {
  (5,5)(6,6)(7,7)(8,8)(9,8)(10,8)(11,10)(12,10)(15,10)(20,10)};
\addlegendentry{Qwen-1.5B}
 
\addplot[black!60,mark=o,thick,mark size=2pt] coordinates {
  (5,5)(6,6)(7,7)(8,8)(9,9)(10,10)(11,11)(12,12)(15,24)(20,24)};
\addlegendentry{Qwen-3B}
 
\addplot[black!30,mark=x,thick,mark size=2pt] coordinates {
  (5,5)(6,6)(7,7)(8,8)(9,9)(10,10)(11,11)(12,12)(15,16)(20,20)};
\addlegendentry{Qwen-7B}
 
\end{axis}
\end{tikzpicture}
 
\caption{Model output vs.\ correct count across sequence lengths for repeated-token counting. The dashed diagonal is the correct answer. Llama-1B collapses to an attractor of ``8'' while Llama-3B collapses to ``14'' after $n{=}10$. Qwen-1.5B stays correct through $n{=}8$ before drifting to ``10''. Qwen-3B and Qwen-7B stay mostly accurate with only late-sequence degradation.}
 
\label{fig:nsweep}
\end{figure}

Figure~\ref{fig:nsweep} shows model outputs across sequence lengths for all five models. The three counting-failure models each settle on a distinct attractor value at a distinct threshold, which is inconsistent with a shared sequence-length estimation heuristic and consistent with model-specific learned priors that activate at specific residual stream states.

Extending the sweep to $n{=}40$ shows the attractor value is stable within the range plotted but is not fixed indefinitely. Beyond $n{\approx}30$ the output drifts to larger values (for example Llama-3.2-3B moves from ``14'' to values above 40 at $n{=}35$--40). The fixed-attractor description therefore applies to the tested range instead of to arbitrary sequence length. 

 
The two models that solve counting show a separate behavioral failure, since Qwen2.5-3B and 7B pass P1 and P3 but fail P2 deterministically. A single banana at position~5, the standard P2 test position, is never detected at either scale. Both models detect a single intruder only at end positions, a recency bias from the last-token position attending to the sequence end. Detection succeeds once one to two intruders are present, with no consistent difference between the two model sizes, so the threshold does not decrease with scale. Edge cases confirm the counting ability itself is intact, since all bananas produces ``0'', all apples produces ``10'' and a single apple produces ``1'' in both models. Appendix~\ref{sec:anomaly_app} and Table~\ref{tab:anomaly} report the position, quantity and token sweeps in full.

\section{Mechanistic Analysis}
\label{sec:mech}
 
\subsection{Attention does not explain the failure}
 
A possible explanation for counting failure is that repeated identical tokens cause attention to collapse onto a few positions, preventing the model from attending to the full list. Table~\ref{tab:attention} evaluates this hypothesis.
 
\begin{table}[h]
\centering
\caption{Attention entropy and uniformity over word-list token positions averaged across all layers. Higher entropy means attention is more spread over the word list. The similar values across P1 and P2 mean attention cannot explain why P1 fails and P2 does not.}
\label{tab:attention}
\small
\setlength{\tabcolsep}{4pt}
\begin{tabular}{lccc}
\toprule
\textbf{Model} & \textbf{P1 entropy} & \textbf{P2 entropy} &
\textbf{P1 uniformity} \\
\midrule
Llama-3.2-1B  & 2.17 & 2.13 & 0.25 \\
Llama-3.2-3B  & 2.14 & 2.12 & 0.23 \\
Qwen2.5-1.5B  & 2.05 & 2.01 & 0.18 \\
Qwen2.5-3B    & 2.04 & 2.03 & 0.18 \\
Qwen2.5-7B    & 2.13 & 2.12 & 0.26 \\
\bottomrule
\end{tabular}
\end{table}
 
The entropy and uniformity values are nearly identical between P1 and P2 across all five models. The beginning-of-sequence (BOS) token dominates the global argmax at every layer in all models across all three phases. When restricted to word-list positions, the argmax collapses to a single apple token for P1 (all positions are identical) and distributes across positions for P3. There is no attention collapse specific to the failing condition in any model. The attention-sink hypothesis is ruled out across all five models.
 
\subsection{The correct count is linearly encoded at every layer}
 
\begin{figure}[h]
\centering
\begin{tikzpicture}
\begin{axis}[
  width=\columnwidth, height=5.0cm,
  xlabel={Layer},
  ylabel={Probe $R^2$},
  xmin=-0.5, xmax=17,
  ymin=-0.35, ymax=1.08,
  xtick={0,2,4,6,8,10,12,14,16},
  xticklabels={E,L2,L4,L6,L8,L10,L12,L14,L16},
  grid=major, grid style={dotted,gray!35},
  legend columns=1,
legend style={
  at={(0.42,-0.25)},
  anchor=north,
  font=\scriptsize,
  inner sep=2pt,
  draw=none,
  column sep=1em,
},
  tick label style={font=\scriptsize},
  label style={font=\small},
]
\addplot[dotted,black,very thick,domain=-0.5:17,samples=2]{0.95};
\node[font=\tiny,black,anchor=west] at (axis cs:14.8,0.92){0.95};
\addplot[cmlp,mark=*,thick,mark size=1.8pt] coordinates {
  (0,-0.2100)(1,0.9985)(2,0.9974)(3,0.9956)(4,0.9926)(5,0.9828)
  (6,0.9833)(7,0.9836)(8,0.9868)(9,0.9924)(10,0.9914)(11,0.9934)
  (12,0.9895)(13,0.9914)(14,0.9910)(15,0.9914)(16,0.9903)};
\addlegendentry{Repeated (P1) --- wrong output}
\addplot[cblue,mark=triangle*,thick,mark size=1.8pt] coordinates {
  (0,-0.2100)(1,0.9856)(2,0.9938)(3,0.9916)(4,0.9830)(5,0.9777)
  (6,0.9786)(7,0.9672)(8,0.9675)(9,0.9615)(10,0.9614)(11,0.9580)
  (12,0.9630)(13,0.9611)(14,0.9581)(15,0.9498)(16,0.9622)};
\addlegendentry{Unique (P3) --- correct output}
\draw[dashed,cmlp,thick]
  (axis cs:14,-0.35)--(axis cs:14,1.08)
  node[above,font=\tiny,cmlp]{L14};
\end{axis}
\end{tikzpicture}
\caption{Linear probe $R^2$ per layer for Llama-3.2-1B. The count is linearly decodable from L01 onward for both conditions. In the mid-to-late layers the repeated-token condition probes more cleanly than the unique-token condition, so the count is encoded more precisely where the model outputs the wrong answer.}
\label{fig:probes}
\end{figure}
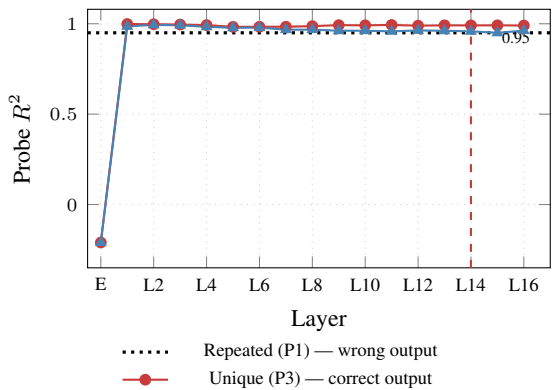

Figure~\ref{fig:probes} shows probe $R^2$ per layer for Llama-3.2-1B. The linear probes achieve $R^2{>}0.99$ from L01 onward for the repeated-token condition. The embedding layer predicts the count only at chance level ($R^2{=}{-}0.17$), indicating that the count information emerges during transformer computation. The same experiment on Llama-3.2-3B yields $R^2{>}0.99$ through the early and middle layers, decaying modestly to 0.93 at the final layer while remaining clearly decodable throughout.

On Qwen2.5-1.5B, $R^2$ ranges from 0.964 to 0.990 (mean 0.978) across the lock-in layers L22--L26. The early layers show weaker probe performance with $R^2{\approx}0.70$--0.90 at L01--L08. On Qwen2.5-3B and 7B, which solve P1, probe $R^2$ peaks at 0.995 (3B) and 0.985 (7B) in the early-to-middle layers and decays modestly to 0.86 (3B) and 0.84 (7B) at the final layer. The count remains clearly decodable throughout in both models and unlike the counting-failure models their output matches the decoded count.
 
The critical finding is the comparison between conditions in Llama-3.2-1B. Repeated-token prompts show lower probe mean absolute error than unique-token prompts in mid-to-late layers: 0.09 to 0.25 for repeated and 0.29 to 0.52 for unique at peak divergence around L08 to L11. The count is encoded more cleanly in the condition where the model outputs the wrong answer than in the condition where it outputs the correct answer. This representation-behavior dissociation rules out encoding failure. The model has the information but something downstream prevents its expression.
 
Table~\ref{tab:logitlens} summarizes logit lens results across all five models. It shows at what depth each model commits to the wrong answer and whether the correct answer ever appears in the output distribution.
 
\begin{table}[h]
\centering
\caption{Logit lens top digit at key layers across all five models (P1, space-separated, normalized depth). Wrong-answer lock-in occurs at 85--93\,\% depth in the failing models. \textit{Numeric from} is the depth at which the logit lens first predicts a digit. \textit{Dashes} mean no wrong-answer lock-in occurs.}
\label{tab:logitlens}
\small
\setlength{\tabcolsep}{3pt}
\begin{tabular}{lccc}
\toprule
\textbf{Model} & \textbf{Numeric from} & \textbf{Lock-in depth} &
\textbf{Wrong ans.} \\
\midrule
Llama-3.2-1B  & 81\,\% (L13) & 87.5\,\% (L14) & ``8''  \\
Llama-3.2-3B  & 79\,\% (L22) & 92.8\,\% (L26) & ``14'' \\
Qwen2.5-1.5B  & 75\,\% (L21) & 85.7\,\% (L24) & ``8''  \\
Qwen2.5-3B    & 75\,\%       & ---             & correct \\
Qwen2.5-7B    & 87.5\,\%     & ---             & correct \\
\bottomrule
\end{tabular}
\end{table}
 
In Llama-3.2-1B, both P1 and P3 produce non-numeric tokens through L12 and enter a numerically meaningful regime at L13. The divergence occurs at L14 (87.5\,\% depth), where P1 commits to ``8'' while P3 resolves to ``10'' by L15. Crucially, ``10'' remains in P1's top-5 at L14 through L16 but is outranked by ``8''. The correct answer is present in the distribution but cannot overcome the prior. In Llama-3.2-3B, P1 and P2 produce identical logit lens projections through L21. The divergence occurs at L22 where P1's top digit becomes ``14'' and P2's remains non-numeric, so the banana suppresses the prior at its point of first formation. In Qwen2.5-1.5B, the wrong answer ``8'' first appears at L22 but does not stabilize until L24 (85.7\,\% depth), which is the primary writer. Qwen2.5-3B and 7B show no wrong-answer lock-in. The correct answer emerges and stabilizes without being overridden. The wrong-answer lock-in across all three failing models emerges at 85--93\,\% network depth showing a consistent pattern across scales.

\subsection{A format-triggered MLP overwrites the correct count}

Table~\ref{tab:mlpdecomp} shows the decomposition and identifies the writer layer for all three models. In every case, the wrong answer is written by a specific MLP block at 85--93\,\% network depth. In Llama-3.2-1B, the MLP at L14 is the sole writer of ``8''. The top digit is ``12'' entering L14, remains ``12'' after the attention sublayer and shifts to ``8'' after the MLP. For P3, the MLP at L16 writes the correct answer ``10'', so the failure and the success are each implemented by a single adjacent MLP block.

\begin{table}[h]
\centering
\caption{MLP vs.\ attention decomposition for all three models where the counting prior is active. Columns $d_0$, $d_1$ and $d_2$ give the top logit-lens digit entering the layer, after the attention sublayer and after the MLP. The Writer column reads off which sublayer changed the digit, where MLP means the MLP writes the wrong answer, MLP\,E means the MLP erases the prior digit, ATT means attention writes the digit and --- means no change. Only layers where the top logit-lens digit changes are shown.}
\label{tab:mlpdecomp}
\small
\setlength{\tabcolsep}{8pt}
\begin{tabular}{llcccc}
\toprule
\textbf{Model} & \textbf{Layer} & $d_0$ & $d_1$ &
$d_2$ & \textbf{Writer} \\
\midrule
\multirow{6}{*}{Llama-1B}
  & L11 & 5  & 5  & 3  & --- \\
  & L12 & 3  & 1  & 12 & --- \\
  & L13 & 12 & 12 & 12 & --- \\
  & \textbf{L14} & \textbf{12} & \textbf{12} & \textbf{8}  & \textbf{MLP} \\
  & L15 & 8  & 8  & 8  & --- \\
  & L16 & 8  & 8  & 8  & --- \\
\midrule
\multirow{8}{*}{Llama-3B}
  & \textbf{L11} & \textbf{17} & \textbf{1}  & \textbf{14} & \textbf{MLP} \\
  & L12 & 14 & 14 & 4  & MLP\,E \\
  & L13 & 4  & 14 & 17 & ATT \\
  & \textbf{L22} & \textbf{8}  & \textbf{8}  & \textbf{14} & \textbf{MLP} \\
  & L23 & 14 & 14 & 14 & --- \\
  & L24 & 14 & 8  & 10 & MLP\,E \\
  & L25 & 10 & 10 & 10 & --- \\
  & \textbf{L26} & \textbf{10} & \textbf{10} & \textbf{14} & \textbf{MLP} \\
\midrule
\multirow{5}{*}{Qwen-1.5B}
  & L21 & 1  & 7  & 4  & --- \\
  & \textbf{L22} & \textbf{4}  & \textbf{4}  & \textbf{8}  & \textbf{MLP} \\
  & L23 & 8  & 7  & 7  & MLP\,E \\
  & \textbf{L24} & \textbf{7}  & \textbf{7}  & \textbf{8}  & \textbf{MLP} \\
  & L25 & 8  & 8  & 8  & --- \\
\bottomrule
\end{tabular}
\end{table}
 
\begin{figure}[H]
\centering
\includegraphics[width=\linewidth, trim= 5 5 5 16, clip]{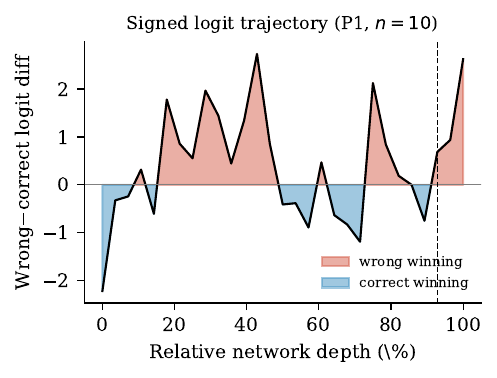}
\caption{Signed logit trajectory for Llama-3.2-3B (P1, $n{=}10$). Wrong-minus-correct logit margin at each layer, shaded red where the wrong answer leads and blue where the correct answer leads. The wrong answer is written and re-written across the late layers, locking in at the L26 writer (dashed).}
\label{fig:trajectory}
\end{figure}

In Llama-3.2-3B the mechanism is distributed across three competing blocks. The full layer sweep reveals ``14'' appears transiently at L11 MLP and is immediately erased at L12, before re-emerging definitively at L22. The MLP at L22 writes ``14'' at exactly $n{=}10$, the threshold where behavioral accuracy collapses. ``14'' persists through L23, then at L24 attention erases it and the layer settles on ``10'' through L25. The MLP at L26 then reinstates ``14'' and stabilizes it through L27--L28. Figure~\ref{fig:trajectory} traces the signed wrong-minus-correct margin across layers, showing the answer written at L22, briefly displaced by ``10'' at L24--L25 and reinstated at the L26 writer.
 
For P2, the MLP at L22 never fires. The banana suppresses the prior where it would otherwise form, which shows up as a logit-lens divergence between P1 and P2 at L22. In Qwen2.5-1.5B, "8" first appears at L22 and is then suppressed at L23, and it does not stabilize until the MLP at L24 reinstates it at 85.7\% relative depth. L24 is the primary writer, where "8" locks in and persists through L28. This two-step pattern mirrors Llama-3.2-3B, where the first write also lands at L22 and a later MLP does the stabilizing.
 
Table~\ref{tab:pern} tests whether the MLP writer is estimating the actual sequence length or responding to a fixed format signature. If the MLP were counting, the digit entering it ($d_0$) would change with $n$. The residual stream entering L14 projects to ``12'' for every $n{=}8$--15 in Llama-3.2-1B and the stream entering L26 encodes ``10'' for $n{=}10$, 11 and 12 in Llama-3.2-3B. The MLP receives the same input and produces the same wrong output regardless of how many tokens are in the list.

\begin{table}[h]
\centering
\caption{Per-$n$ decomposition at the writer layer for all three models. As in Table~\ref{tab:mlpdecomp}, $d_0$, $d_1$ and $d_2$ are the top logit-lens digit entering the writer layer, after the attention sublayer and after the MLP. Bold marks the wrong answer being written. If $d_0$ tracked the count it would vary with $n$, which it does not.}
\label{tab:pern}
\small
\setlength{\tabcolsep}{12pt}
\begin{tabular}{llccc}
\toprule
\textbf{Model} & $n$ & $d_0$ & $d_1$ &
$d_2$ \\
\midrule
\multirow{6}{*}{Llama-1B (L14)}
  & 7  & 12 & 5  & 5  \\
  & 8  & 12 & 12 & \textbf{8}  \\
  & 9  & 12 & 12 & \textbf{8}  \\
  & 10 & 12 & 12 & \textbf{8}  \\
  & 12 & 12 & 12 & \textbf{8}  \\
  & 15 & 12 & 12 & \textbf{8}  \\
\midrule
\multirow{5}{*}{Llama-3B (L26)}
  & 7  & 6  & 6  & 7  \\
  & 9  & 8  & 8  & 8  \\
  & 10 & 10 & 10 & \textbf{14} \\
  & 11 & 10 & 10 & \textbf{14} \\
  & 12 & 10 & 10 & \textbf{14} \\
\midrule
\multirow{3}{*}{Qwen-1.5B (L23)}
  & 7  & 7  & 7  & 7  \\
  & 8  & 8  & 7  & 7  \\
  & 10 & 8  & 7  & 7  \\
\bottomrule
\end{tabular}
\end{table}
 
Table~\ref{tab:ablation} asks whether removing L14 fixes the failure. Zeroing L14 shifts the wrong answer from ``8'' to ``16'' for $n{\geq}12$ and leaves the output unchanged at $n{=}9$--11, so L14 is causally involved but the correct answer ``10'' never emerges. The prior is partially distributed, since L14 is a necessary component whose removal only redirects the failure and does not correct it.

\begin{table}[H]
\centering
\caption{Zero-ablation of L14 MLP in Llama-3.2-1B. Normal output has L14 intact. L14 ablated replaces the L14 MLP output with zeros. Fixed means the ablated output equals the correct answer (10). No condition is fixed, since removing L14 shifts the wrong answer but never recovers the correct one.}
\label{tab:ablation}
\small
\setlength{\tabcolsep}{6pt}
\begin{tabular}{lccc}
\toprule
$n$ & \textbf{Normal output} & \textbf{L14 ablated} & \textbf{Fixed?} \\
\midrule
8  & 8 & 7  & \textcolor{cwrong}{no} \\
9  & 8 & 8  & \textcolor{cwrong}{no} \\
10 & 8 & 8  & \textcolor{cwrong}{no} \\
11 & 8 & 8  & \textcolor{cwrong}{no} \\
12 & 8 & 16 & \textcolor{cwrong}{no (shifts to 16)} \\
15 & 8 & 16 & \textcolor{cwrong}{no (shifts to 16)} \\
\bottomrule
\end{tabular}
\end{table}

Table~\ref{tab:ablation3b} reports the analogous interventions at the L26 writer in Llama-3.2-3B, scored by movement in the correct-minus-wrong logit difference instead of by whether the output flips. Both zero- and mean-ablation move the output toward the correct answer at every $n$. The mean-ablation produces the larger effect (movement 2.6--5.9 for $n$ below 15) and shifts the output off the ``14'' attractor at every $n$ below 15. The writer is therefore causally decisive in the larger model as well and the graded logit-difference movement makes this visible where a strict output-flip criterion would not.
 
\begin{table}[h]
\centering
\caption{Ablation of the L26 MLP writer in Llama-3.2-3B. out is the ablated output and $\Delta$ is the movement in the correct-minus-wrong logit difference, where higher is more toward correct. Zero-ablation zeros the L26 MLP output and mean-ablation replaces it with its mean over the P3 condition.}
\label{tab:ablation3b}
\small
\setlength{\tabcolsep}{15pt}
\begin{tabular}{lcccc}
\toprule
 & \multicolumn{2}{c}{\textbf{Zero-ablation}} & \multicolumn{2}{c}{\textbf{Mean-ablation}} \\
\cmidrule(lr){2-3}\cmidrule(lr){4-5}
$n$ & out & $\Delta$ & out & $\Delta$ \\
\midrule
8  & 8  & 1.75 & 8  & 2.62 \\
9  & 8  & 2.38 & 9  & 5.88 \\
10 & 14 & 2.25 & 11 & 3.75 \\
11 & 10 & 4.12 & 11 & 5.38 \\
12 & 14 & 5.00 & 13 & 5.88 \\
15 & 14 & 1.50 & 14 & 1.88 \\
\bottomrule
\end{tabular}
\end{table}

\subsection{The trigger is format- and word-type-specific}
 
The prior survives most surface rewordings of the instruction. Across five paraphrases of the P1 instruction, the residual stream state entering L26 differs between the phrasings that fire the prior and those that do not, which places the trigger upstream of the writer. For \texttt{original} and \texttt{how\_many} the state entering L26 is ``10'' in both cases, yet L26 fires only for \texttt{original}, so the value entering the writer does not by itself determine whether the prior fires. The comma-separated format acts the same way, altering the upstream state so that the activation pattern that triggers the prior never arrives at L26. Appendix~\ref{sec:paraphrase_app} gives the per-paraphrase values.

Substituting the repeated word while holding the task framing fixed isolates which token types trigger the prior. The prior fires for every alphabetic and symbolic token tested and does not fire for digit tokens, which produce the correct count under the same instruction. The prior is therefore word-type-specific and is a learned association between the repeated-word-list format and a specific output value. Appendix~\ref{sec:symbol_app} gives the full substitution set.

The same upstream mechanism predicts that any delimiter breaking the contiguous-space format should disengage the prior instead of the comma alone. We test this by inserting newline or pipe delimiters between items while holding the item count fixed. In Llama-3.2-3B every delimiter tested restores correct counting and the L26 MLP contribution to the wrong-minus-correct logit difference reverses sign from $+1.75$ under the space format to between $-2.1$ and $-4.1$ under each delimiter. In Llama-3.2-1B and Qwen2.5-1.5B no delimiter restores the correct output, matching their behavior under comma format. The trigger is therefore tied to the contiguous-space repeated-token format and is disengaged at the writer layer. Appendix~\ref{sec:interleaved_app} reports the full variant set.

\subsection{The routing failure holds across model families}
 
Table~\ref{tab:crossfamily} summarizes the cross-family mechanistic comparison across the three models where the counting prior is active. The MLP writer appears at 85--93\,\% relative depth despite different architectures and training data.
 
\begin{table}[H]
\centering
\caption{Cross-family mechanistic comparison. Depth is the relative position of the primary MLP writer. $R^2$ at lock-in is measured at the writer layer.}
\label{tab:crossfamily}
\small
\begin{tabular}{lccc}
\toprule
\textbf{Model} & \textbf{Writer} & \textbf{Depth} &
\textbf{$R^2$ at lock-in} \\
\midrule
Llama-3.2-1B  & L14      & 87.5\,\%       & 0.997   \\
Llama-3.2-3B  & L22, L26 & 78.5--92.8\,\% & 0.98, 0.95 \\
Qwen2.5-1.5B  & L24      & 85.7\,\%       & 0.977   \\
\bottomrule
\end{tabular}
\end{table}

\begin{figure}[h]
\centering
\includegraphics[width=\linewidth, trim=5 5 5 16, clip]{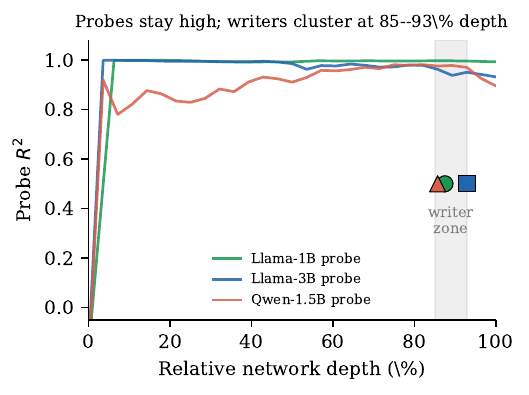}
\caption{The MLP writer sits at 85--93\,\% relative depth across all three counting-failure models. The curves give probe $R^2$ per layer and the markers give each model's writer layer. Despite different architectures and parameter counts, the writer falls in a narrow late-network band (shaded).}
\label{fig:crossfamily}
\end{figure}

The probe dissociation holds at the lock-in layers in all three models. This depth regularity and the probe dissociation together establish that the routing failure is not an artifact of any architecture or training recipe. Figure~\ref{fig:crossfamily} makes the regularity visual, overlaying the three models on a shared relative-depth axis with probe $R^2$ staying high up to each writer.
 
\begin{figure}[h]
\centering
\begin{tikzpicture}
\begin{axis}[
  width=\columnwidth, height=4.5cm,
  xlabel={Layer},
  ylabel={Banana / apple attention ratio},
  xmin=0, xmax=37, ymin=0, ymax=4.5,
  grid=major, grid style={dotted,gray!35},
  tick label style={font=\scriptsize},
  label style={font=\small},
]
\addplot[cblue,mark=*,thick,mark size=1.2pt] coordinates {
  (1,0.945)(2,0.838)(3,1.079)(4,0.386)(5,2.601)(6,0.931)(7,2.117)
  (8,2.652)(9,0.957)(10,0.980)(11,1.841)(12,2.385)(13,0.697)(14,1.180)
  (15,1.332)(16,1.921)(17,2.359)(18,4.109)(19,1.269)(20,2.134)(21,0.963)
  (22,0.921)(23,0.619)(24,0.953)(25,0.842)(26,0.539)(27,0.807)(28,0.364)
  (29,0.915)(30,0.636)(31,0.662)(32,0.878)(33,0.962)(34,0.757)(35,1.704)
  (36,1.469)};
\addplot[dashed,gray,thick,domain=0:37,samples=2]{1};
\node[font=\tiny,gray] at (axis cs:34,1.12){uniform};
\addplot[dashed,cmlp,domain=0:37,samples=2]{1.5};
\node[font=\tiny,cmlp] at (axis cs:33.5,1.62){over-attended};
\end{axis}
\end{tikzpicture}
\caption{Per-layer ratio of banana to mean apple attention (Qwen2.5-3B, P2, last token, averaged over heads and renormalized to word-list positions). The banana is over-attended in 10 of 36 layers at two to four times the apple attention, yet the model outputs the wrong count, which places the failure at the attention-to-output stage.}
\label{fig:bananaattn}
\end{figure}
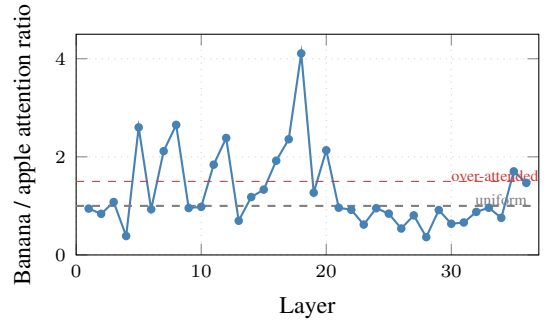

Qwen2.5-3B solves counting but fails anomaly detection. Figure~\ref{fig:bananaattn} shows that the banana is over-attended (ratio~${>}1.5$) in 10 of 36 layers, receiving 2--4$\times$ the attention of individual apple tokens. Per-head analysis at the most banana-ignoring layers shows most heads suppress banana attention at late layers (L28), with one head (H11 at L28) strongly over-attending the banana (ratio 5.9) while the mean collapses to 0.36. Entropy comparison shows P2 attention is more concentrated than P1 at L18 ($\Delta H = -0.16$) and L34 ($\Delta H = -0.24$), consistent with the model committing to a wrong answer and attention collapsing onto apple positions. The model detects the anomaly at the attention level but does not convert this detection into a corrected count. This is a routing failure at the attention-to-output level, parallel to the residual-stream routing failure in smaller models but at a different representational stage. In both cases the relevant information is present inside the model and a specific computation fails to express it.
 
\section{Discussion}
\label{sec:discussion}
 
\paragraph{Representation versus routing.}
The central finding of this paper is a dissociation. The correct count is linearly encoded in the residual stream from the first transformer layer, yet a format-triggered MLP block at roughly 85--93\,\% network depth overwrites
it with a fixed wrong answer. Counting failure in these models is not a failure to represent the count. It is a failure to route the correctly-encoded count to the output.
 
\paragraph{The format trigger.}
The format-trigger mechanism provides a circuit-level account of the behavioral findings. In Llama-3.2-3B, comma-separated format alters the upstream residual stream state at L26 so that the prior does not receive the activation signature that fires it and the count then passes through unchanged. Paraphrase variations that produce a different state at the writer layer also
bypass the prior. In Llama-3.2-1B, comma format does not alter the state at L14 enough to suppress the prior. The ``8'' attractor persists across both formats. The scale boundary for format sensitivity is mechanistically grounded in whether the format change propagates to the writer layer in a form the prior pattern-matches.
 
\paragraph{Word-type specificity.}
The word-versus-digit dissociation adds precision to the prior's scope. It fires for alphabetic word-tokens but not digit-tokens, indicating it is a learned association between the repeated-word-list format and a specific output value rather than a general sequence-length estimator. The same model that fails to count repeated words counts repeated digits correctly under identical task framing. The failure is circuit-specific.
 
\paragraph{Routing failures across scales.}
Routing failures appear at different representational levels across both families and failure modes. In smaller models the residual stream encodes the correct count but an MLP overwrites it. In larger Qwen models that solve counting, the attention mechanism detects the
anomaly but the detection signal is not converted to a corrected count. These are structurally the same type of failure. Information is present and a specific computation fails to use it. Behavioral failures in capable models may more often be failures of expression
than failures of representation and mechanistic analysis of the routing pathway is necessary to understand them.
 
\section{Limitations}
\label{sec:limitations}
 
Mechanistic analysis was conducted on single prompts per condition and while paraphrase experiments cover five surface-form variations, systematic analysis across a larger prompt distribution would strengthen the generalization claims. The word-versus-digit dissociation was tested only on Llama-3.2-1B and may not generalize across model families. All models tested are instruction-tuned; base model behavior is not analyzed.
The Qwen2.5 logit lens is limited to single-digit tokens due to the tokenizer, which prevents direct tracking of the ``10'' correct answer trajectory.
 
\section{Conclusion}
\label{sec:conclusion}
 
Mechanistic analysis of a simple counting task reveals that behavioral failures and internal representations can dissociate completely. We applied linear probing, logit lens projection and MLP-level decomposition
across five models and two families to locate the specific computation responsible for the failure. The failure is not in the representation but in a single MLP block that fires on a format signature and overwrites a correctly-encoded answer with a fixed wrong one. The same methodology applied to larger models that solve counting uncovers a second routing failure at the attention level, where anomaly detection signals are present but not expressed. Both cases show that a model can have the right answer internally and still
produce the wrong output. Identifying which type of failure is present is the necessary first step
before any intervention can succeed.

\section*{Impact Statement}
This work studies why large language models fail at a simple counting task by locating the internal computation responsible for the failure. We show that the failure is a routing failure and that standard mechanistic interpretability  tools are sufficient to locate and characterize it. All experiments were conducted on publicly released models available through Hugging Face. The authors have followed the terms of use 
and licensing agreements of all models and software libraries used in this research.

\bibliography{custom}

\clearpage
\appendix

\section{Additional Experiments}
\label{sec:additional}
 
\subsection{Anomaly detection}
\label{sec:anomaly_app}
 
Qwen2.5-3B and 7B solve repeated-token counting but fail to detect a single intruder token, as reported in Section~\ref{sec:behavioral}. We characterize that failure by varying the intruder position, the number of intruders and the intruder token. With a single banana intruder at $n{=}10$, both models produce a non-10 output only at positions 6, 7 and 9. Neither model detects the intruder at position 5, which is the position used in the P2 condition. Detection succeeds once the number of intruders increases. Banana, car and seven are detected at two intruders and xyz is detected at one. Varying the base length gives thresholds of three at $n{=}8$, two at $n{=}10$ and one at $n{=}12$. The thresholds are close enough across the two model sizes that we draw no scaling conclusion. The position-level behavior is irregular, so the quantity threshold is the more stable measurement.
 
\begin{table}[H]
\centering
\caption{Anomaly detection for Qwen2.5-3B and 7B. Detected positions are list indices where a single banana intruder at $n{=}10$ triggers a non-10 output. Min. intruders is the fewest same-type intruders needed to produce a non-10 output, across the four intruder tokens tested. Both models fail with a single intruder and detect at a low threshold that does not decrease with
scale.}
\label{tab:anomaly}
\small
\setlength{\tabcolsep}{4pt}
\begin{tabular}{lccc}
\toprule
\textbf{Model} & \textbf{Detected pos.} & \textbf{Min. intruders} &
\textbf{P2 accuracy} \\
\midrule
Qwen-3B & 6, 7, 9 & 1--2 & \textcolor{cwrong}{0\,\%} \\
Qwen-7B & 6, 7, 9 & 1--2 & \textcolor{cwrong}{0\,\%} \\
\bottomrule
\end{tabular}
\end{table}
 
\subsection{Paraphrase robustness}
\label{sec:paraphrase_app}
 
The counting prior survives most surface rewordings of the instruction. We test five paraphrases of the P1 instruction on Llama-3.2-3B. For each paraphrase we record the behavioral output and the residual stream state entering the writer layer. The \texttt{original} phrasing produces the attractor 14. The \texttt{list\_first} and \texttt{simple} phrasings produce 8 and the \texttt{how\_many} and \texttt{tally} phrasings produce the correct answer 10. The state entering L26 is 10 for \texttt{original} and \texttt{how\_many} and 8 for the remaining three. The same input value therefore reaches the writer under a phrasing that fires the prior and under one that does not. This places the trigger upstream of L26. The L26 MLP contribution is $+1.75$ for \texttt{original} and $+1.06$ for \texttt{how\_many}. The sign of the contribution does not separate the two cases, so we report per-paraphrase values in place of a single mechanism.
 
\begin{table}[H]
\centering
\caption{Paraphrase mechanistic analysis for Llama-3.2-3B. As in Table~\ref{tab:mlpdecomp}, $d_0$ and $d_2$ are the top logit-lens digit entering the layer and after the MLP, shown here at the two diagnostic layers L11 and L26. Diagnostic~1 shows which paraphrases cause L11 MLP to write ``14'' transiently. Diagnostic~2 shows the residual stream state entering L26 per paraphrase. L26 fires only when it receives ``10'' as input.}
\label{tab:paraphrase_mech}
\small
\setlength{\tabcolsep}{8pt}
\begin{tabular}{lccccl}
\toprule
 & \multicolumn{2}{c}{\textbf{L11}} & \multicolumn{2}{c}{\textbf{L26}} & \\
\cmidrule(lr){2-3}\cmidrule(lr){4-5}
\textbf{Paraphrase} & $d_0$ & $d_2$ & $d_0$ & $d_2$ & \textbf{Output} \\
\midrule
original   & 17 & \textbf{14} & 10 & \textbf{14} & 14\,\textcolor{cwrong}{$\times$} \\
how\_many  & 18 & 17          & 10 & 10          & 10\,\textcolor{cright}{$\checkmark$} \\
list\_first & 18 & \textbf{14} &  8 &  8          & 8\,\textcolor{cwrong}{$\times$} \\
tally      & 19 &  5          &  8 &  8          & 10\,\textcolor{cright}{$\checkmark$} \\
simple     & 17 & \textbf{14} &  8 &  8          & 8\,\textcolor{cwrong}{$\times$} \\
\bottomrule
\end{tabular}
\end{table}

\subsection{Symbol substitution}
\label{sec:symbol_app}
 
Replacing the repeated word while holding the task framing fixed isolates which token types trigger the prior. On Llama-3.2-1B at $n{=}10$, the tokens apple, cat, the, a and X all produce the attractor 8. The digit tokens 1 and 0 produce the correct answer 10 and the digit token 7 produces 9. The prior fires for every alphabetic and symbolic token tested and remains silent for digit tokens. This supports the reading that the trigger is a learned association with the repeated-word-list format instead of a general sequence-length estimator.
 
\begin{table}[H]
\centering
\caption{Word vs.\ digit symbol substitution (Llama-3.2-1B, $n{=}10$). The prior fires for all alphabetic and symbolic tokens but not for digit-tokens, which produce the correct answer.}
\label{tab:symbols}
\small
\setlength{\tabcolsep}{4pt}
\begin{tabular}{lcc}
\toprule
\textbf{Symbol} & \textbf{Output} & \textbf{Prior fires?} \\
\midrule
apple & 8  & \textcolor{cwrong}{YES} \\
cat   & 8  & \textcolor{cwrong}{YES} \\
the   & 8  & \textcolor{cwrong}{YES} \\
a     & 8  & \textcolor{cwrong}{YES} \\
X     & 8  & \textcolor{cwrong}{YES} \\
\midrule
1     & 10 & \textcolor{cright}{no} \\
0     & 10 & \textcolor{cright}{no} \\
7     & 9  & \textcolor{cright}{no} \\
\bottomrule
\end{tabular}
\end{table}
 
\subsection{Interleaved structural noise}
\label{sec:interleaved_app}
 
The comma-separated condition in the main text changes both the delimiter and the instruction wording. To separate these, we insert newline and pipe delimiters between items in two forms. One form leaves the instruction unchanged and the other names the delimiter. Every variant holds the item count fixed at ten. On Llama-3.2-3B the space format produces the attractor 14 across all ten seeds. All delimiter variants yield the correct value of 10, hence improving the accuracy level from zero to one. The L26 MLP contribution to the wrong-minus-correct logit difference is $+1.75$ under the space format. It falls to $-4.13$ under comma, $-2.13$ under newline and $-3.13$ under pipe. On Llama-3.2-1B and Qwen2.5-1.5B no delimiter restores the correct output, matching their behavior under comma format.
 
\begin{figure}[h]
\centering
\includegraphics[width=\linewidth, trim=5 5 5 16, clip]{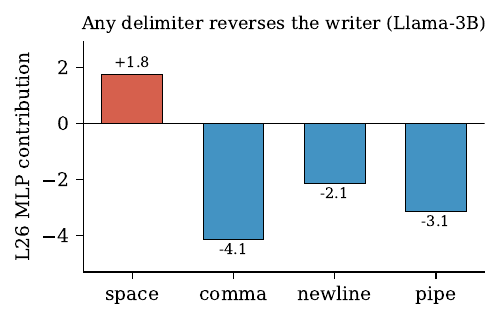}
\caption{Any delimiter that breaks the contiguous-space format reverses the writer (Llama-3.2-3B). The L26 MLP contribution to the wrong-minus-correct logit difference is positive under the space format (writer pushes toward the wrong answer) and negative under comma, newline and pipe delimiters (writer disengaged).}
\label{fig:delimiter}
\end{figure}

\section{Prompt Templates}
\label{sec:prompts}
 
All prompts are reproduced verbatim as sent to the models, with \texttt{\{list\}} denoting the item sequence where a template takes one.
 
\begin{tcolorbox}[prompttitled, title={Three phase conditions, space-separated}]
\textbf{P1 Baseline} (expected 10)\\
Count the number of times "apple" appears in this list: apple apple apple apple apple apple apple apple apple apple. Respond only with the integer, nothing else.\par\smallskip
\textbf{P2 Anomaly} (expected 9)\\
Count the number of times "apple" appears in this list: apple apple apple apple banana apple apple apple apple apple. Respond only with the integer, nothing else.\par\smallskip
\textbf{P3 Control} (expected 10)\\
Count the number of words in this list: dog cat car red blue green house tree book pen. Respond only with the integer, nothing else.
\end{tcolorbox}
 
\begin{tcolorbox}[prompttitled, title={Comma-separated variants}]
Identical to the three conditions above, with \textnormal{\emph{in this comma-separated list:}} in the instruction and the items joined by ", ".
\end{tcolorbox}
 
Each delimiter in the interleaved-noise set appears in a no-hint form, where the instruction is unchanged, and a hint form, where the instruction names the delimiter. This separates the format shift from the instruction wording, which the comma condition bundles together.
 
\begin{tcolorbox}[prompttitled, title={Interleaved-noise variants, P1 only, $n{=}10$}]
\textbf{newline\_no\_hint}\quad items joined by " \textbackslash n ", instruction reads \emph{in this list:}\par\smallskip
\textbf{newline\_hint}\quad items joined by " \textbackslash n ", instruction reads \emph{in this newline-separated list:}\par\smallskip
\textbf{pipe\_no\_hint}\quad items joined by " | ", instruction reads \emph{in this list:}\par\smallskip
\textbf{pipe\_hint}\quad items joined by " | ", instruction reads \emph{in this pipe-separated list:}
\end{tcolorbox}
 
\begin{tcolorbox}[prompttitled, title={Five paraphrases of the P1 instruction}]
\textbf{original}\\
Count the number of times "apple" appears in this list: \{list\}. Respond only with the integer, nothing else.\par\smallskip
\textbf{how\_many}\\
How many times does the word "apple" appear in the following list: \{list\}? Answer with a single integer, nothing else.\par\smallskip
\textbf{list\_first}\\
List: \{list\}\textbackslash nHow many times does "apple" appear? Single integer only.\par\smallskip
\textbf{tally}\\
Tally the occurrences of "apple" in this sequence: \{list\}. Output only the count as an integer.\par\smallskip
\textbf{simple}\\
Count "apple" in: \{list\}. Reply with just the number.
\end{tcolorbox}
 
\begin{tcolorbox}[prompttitled, title={Chain-of-thought variants}]
All share the stem \emph{Count the number of times "apple" appears in this list: \{apples\}.} followed by\par\smallskip
\textbf{direct}\quad Respond only with the integer, nothing else.\par\smallskip
\textbf{cot\_stepbystep}\quad Think step by step, then give the final count as an integer on the last line.\par\smallskip
\textbf{cot\_enumerate}\quad Number each item in the list, then give the total count on the last line.\par\smallskip
\textbf{cot\_tally}\quad Go through each word one by one, keep a running tally, then output only the final number.
\end{tcolorbox}
 
\begin{tcolorbox}[prompttitled, title={Language variants}]
\textbf{English}\quad the P1 template above.\par\smallskip
\textbf{Chinese}\quad \begin{CJK}{UTF8}{gbsn}请计算"apple"在以下列表中出现的次数：apple apple apple apple apple apple apple apple apple apple。只回答数字，不要其他内容。\end{CJK}\par\smallskip
\textbf{French}\quad Comptez le nombre de fois que "apple" appara\^{i}t dans cette liste: apple apple apple apple apple apple apple apple apple apple. R\'{e}pondez uniquement avec le nombre entier.
\end{tcolorbox}
 
\begin{tcolorbox}[prompttitled, title={Token sets}]
\textbf{Symbol substitution}\quad ["apple", "1", "0", "7", "cat", "the", "a", "X"], each substituted for the repeated token in the P1 template.\par\smallskip
\textbf{Unique-word vocabulary}\quad dog, cat, car, red, blue, green, house, tree, book, pen, fish, cup, hat, sun, moon, sky, fire, rain, snow, wind. Used for P3 and for the unique-token condition of the $n$-sweep, taking the first $n$ words.
\end{tcolorbox}
 
\section{Reproducibility}
\label{sec:reprod}
 
All five models are instruction-tuned checkpoints obtained from HuggingFace. All experiments ran on a single NVIDIA L40S GPU provisioned through RunPod. The model weights were loaded in bfloat16 with automatic device mapping.
 
The maximum number of new tokens was 16 for the counting conditions, 8 for the sequence-length sweeps and 300 for the chain-of-thought conditions. The precision and seed robustness check repeats the writer-layer localization in both bfloat16 and float32 across seeds 42, 123 and 456.
 
The linear probes use ridge regression with a regularization strength of 10.0 on standardized features, evaluated by leave-one-out cross-validation. The probe training sets use $n$ from 3 to 15 for the repeated-token condition and $n$ from 3 to 13 for the unique-token condition, taking the last-token residual stream activation at every layer.

\subsection{Disclosure of AI Assistance}
\label{sec:disclosure}
Gemini was used for grammar correction and sentence-level polish during the writing of this paper. The research idea, experimental design, results and their interpretation are the sole work of the authors.

\end{document}